\documentclass[10pt, conference, compsocconf]{IEEEtran}
\usepackage{graphicx}

\ifCLASSINFOpdf
\else
\fi
\begin{document}
%
\title{Centroid Based Binary Tree Structured SVM for Multi Classification}


\author {\IEEEauthorblockN{Aruna Govada, Bhavul Gauri, S.K.Sahay }
\IEEEauthorblockA{BITS-Pilani, KK Birla Goa Campus: CSIS\\
Goa, India\\
Email: garuna@goa.bits-pilani.ac.in, bhavul93@gmail.com, ssahay@goa.bits-pilani.ac.in}
}


%


\maketitle

\begin{abstract}
Support Vector Machines (SVMs) were primarily designed for 2-class classification. But they have been extended for N-class classification also based on the requirement of multiclasses in the practical applications. Although  N-class classification using SVM has considerable research attention, getting minimum number of classifiers at the time of training and testing is still a continuing research. We propose a new algorithm CBTS-SVM (Centroid based Binary Tree Structured SVM) which addresses this issue. In this we build a binary tree of SVM models based on the similarity of the class labels by finding their distance from the corresponding centroids at the root level. The experimental results demonstrates the comparable accuracy for CBTS  with OVO with reasonable gamma and cost values. On the other hand when CBTS is compared with OVA, it gives the better accuracy with reduced training time and testing time. Furthermore CBTS is also scalable as it is able to handle the large data sets. 
\end{abstract}


%
\IEEEpeerreviewmaketitle

\section{Introduction}
\par A Support Vector Machine (SVM) is a distinguishing classifier conventionally defined by a separating hyperplane. In other words, given labeled training data (supervised learning), the algorithm outputs an optimal hyperplane which classifies the unseen examples[1-2]. SVM is not limited statistics or machine learning but can be applied in wide number of applications. SVM proved to be the best classifier in different applications from handwritten digit recognition to text categorization. SVM doesn't have any effect on classification due to the curse of dimensionality. It works well with high dimensional data.
\par Literature survey witnesses that the Support vector machines are the best classifiers for 2-class classification problems. The real time applications are not limited to binary classification but multiclass classification[3]. For example, to classify an astronomical object as a star, a galaxy or a quasar requires a multi-class classifier but not binary classifier. 
\par There are two major approaches in solving the N-class problem one is a single large optimization problem [4-5], and the alternative is to decompose N-class problem into multiple 2-class problems. But solving a single large optimization problem will be expensive in terms of computational time and not suitable for practical applications. There are several algorithms based on the second approach like OVO(one-versus-one),OVA(one-versus-all), DAG(Directed Acyclic Graph)[6-7]. In this paper we propose a new algorithm which decomposes the single N-class problem into multiple 2-class problems. And it requires less number of binary classifiers when compared with the above mentioned algorithms and gives a better accuracy too.

\par The rest of the paper is organized as follows. The related work is discussed in section 2.The proposed approach is presented in section 3. Section 4 contains the experimental results and comparison of the proposed approach with the existing algorithms. Finally, section 5 contains Conclusion and future work.


\section{Related Work}
\subsection {Binary SVM}

The learning task in binary SVM can be represented as the following
$$min_w = \frac{\parallel w \parallel^2}{2}$$

subject to $y_i(w.x_i + b) \ge 1, \quad i =1,2,....k$  where $w$ and $b$ are the parameters of the model for total $k$ number of instances.\\

 Using Lagrange multiplier method the following equation is to be solved,

              $$ L_p = \frac{\parallel w \parallel^2}{2} - \sum_{i=1...k} \lambda_i (y_i(w.x_i   + b) -1)$$ \\ 
            The dual version of the above problem is \\
    $$ L_D = \sum_{i=1...k} \lambda_i - \frac{1}{2} \sum_{i,j} \lambda_i \lambda_j y_iy_jx_i.x_j $$ 
    subject to $$\lambda_i \ge 0 $$
     $$ \lambda_i (y_i(w.x_i   + b) -1)=0 $$ 
where $\lambda_i $ are known as the Lagrange multipliers.\\
By solving this dual problem, SVM will be found. Once the SVM model is built,the class label of a testing object z can be predicted as follows.\\
   $$  f(z) = sign\sum_{i=1...k} (\lambda_i y_i x_i.z + b) $$ 
   if $ f(z) \ge 0 $ z will be predicted as + class else -ve class.
   
\subsection {Multi-Class SVM}

\subsubsection {\textbf {One versus All}}
The simple approach is to decompose the problem of classifying N classes into N binary problems, where each problem differentiates a given class from the other N-1 classes [8]. In this approach, we require K=N binary classifiers, where the Nth classifier is trained with positive examples belonging to class N and negative examples belonging to the other N-1 classes. When an unknown example is to be predicted, the classifier achieving the maximum output is considered as the best choice, and the corresponding class label is assigned to that test object. Though this approach is simple [8], it provides performance that is comparable to other more complicated approaches when the binary classifier is tuned well.

\subsubsection{\textbf {One versus One}}
In this approach, each class is compared to every other class [9-10]. A binary classifier is built to differentiate between each pair of classes, while discarding the rest of the classes. This requires building N(N-1)/2 binary classifiers. When testing a new object, a voting is performed among the classifiers and the class with the maximum number of votes will be considered as the best choice. Results [6,11]show that this approach is in general better than the one-versus-all approach.

\section{The Proposed Approach}
The N-class problem is decomposed into multiple 2-class problems in a binary tree structured manner. K-Means clustering is used as a preprocessing step to get the rough estimation of similarity between the class labels. This will let us divide the class labels into two disjoint sets and build the SVM for the root node. Thereafter every node is divided at the mid point for creating disjoint sets. The order of the class labels is computed based on the SSE. The least SSE will be first in the list and the highest SSE will appear last in the list.
This way (n-1) binary SVMs will be built, and hence needs only (n-1)/2 SVMs evaluation to classify the unclassified record. This is better than the worst case (n-1) of OVA and n(n-1)/2 of OVO. At the same time, our experimental results show that the accuracy is comparable with OVO. But when CBTS is compared with OVA it shows a better accuracy with reduced training time and testing time. The algorithm for training and testing is illustrated below.

\subsection
{The Training model of CBTS SVM } 
Input the training objects. Add all the training objects to the root node. Let the class labels are from 1....N \\
\textbf{Preprocessing Step:} Divide the training objects i.e.the root node into two clusters/nodes ($I_L$) and ($I_R$) using K-Means clustering (Centroid based). \\
\begin{enumerate}
\item The objects will be adjusted to ($I_L$) as positive class or ($I_R$) as negative class based on the majority of their class labels from the two clusters. \\
\item For ($I_L$) and ($I_R$) calculate the SSE of all objects based on the same class labels  and sort them in the ascending order. The SSE is given by  $ SSE = \sum_{i=1...k}  (x_{i},C)$ \\
\item For both ($I_L$) and ($I_R$) Repeat \\
\begin {enumerate}
\item If the number of class labels of the node are two, construct the binary classifier and return. If the number of class labels are more than two \\
\item Divide the each node exactly at the mid point, construct the binary SVM and repeat this till we reach only two class labels.
\end {enumerate}
\end {enumerate}
\subsection
{The Testing model of CBTS SVM }
\begin{enumerate}
\item The test object should be evaluated on the root node the binary tree of SVMs.\\
\item Repeat: 
\begin{itemize}
\item If the value is positive traverse to the left node ($I_L$) else to the right node ($I_R$).\\
\item Until we reach the leaf node.\\
\end{itemize}
\item Classify the test object into the class label of leaf node.	\\
\end{enumerate}

\textbf{Let's run through an example.} Consider figure 1, suppose if we have 8 classes, we first run  k-means clustering with k=2 to divide the data objects according to their distribution. Then, through the cluster distribution and based on the majority we get to know which class labels fall on one side (positive) and which ones on the other side (negative). In the example shown, set $I_L$ \{1,3,7\} belong to the positive class and set of rest of class labels, $I_R$ \{4,6,8,2,5\} belong to the negative class. The order of class labels within the node will be in the ascending order of SSE. We then build a SVM model by constructing a binary SVM between data objects belonging to $I_L$ as belonging to positive class and those belonging to $I_R$ as negative class. 

\par Now, for the left child of the root node, sample space is all the data objects that belong to class labels in $I_L$  i.e. \{1,3,7\}. We divide exactly at the mid point (Remember, clustering is done at the first step only!) and hence make new $I_L$ and $I_R$ sets for this node. So, new $I_L$ is \{1,3\} and $I_R$ is \{7\}. We now consider all data objects having class labels in $I_L$ as having positive class and all in $I_R$ as negative class and build a SVM model. This SVM model acts as left child of the root node. 
\par This way we build the whole Binary Tree of SVM functions recursively for both left and right nodes till we reach a leaf node. When an unseen object has to be classified, the search starts from the root node and then it moves on to the left or right based on the evaluation function value recursively till the leaf node and assigns the corresponding class label to the test object.

\begin{figure}
\centering
\includegraphics[scale=0.42]{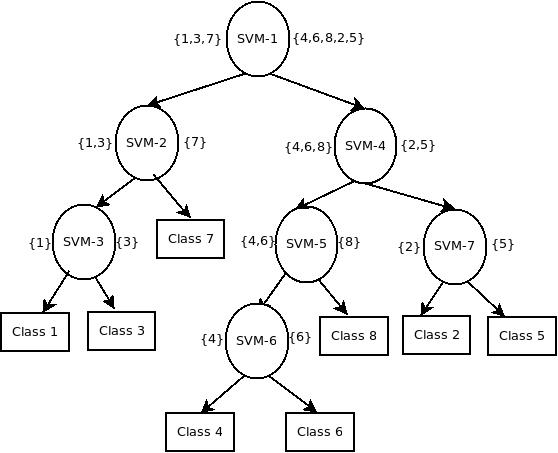}
\caption{The Binary Tree Structured SVM.}
\label{fig:ex}
\end{figure}

 \begin{figure}
\includegraphics[scale=0.42]{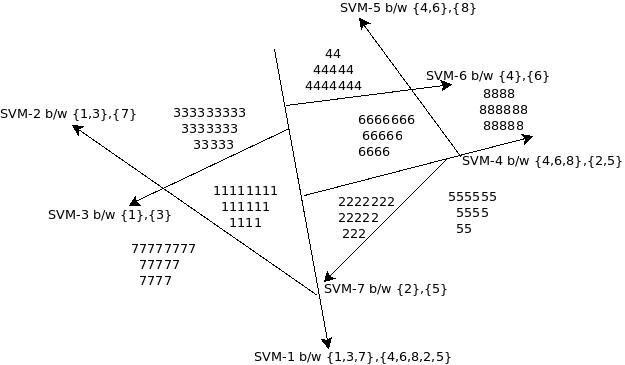}
\caption{The Binary Tree Structured SVM.}
\label{fig:ex}
\end{figure}
In Figure 2 ,the hyperplanes are shown for different class labels based on the above approach. The order of the construction of the hyperplanes will be based on the binary tree structure.

\section {Experimental Analysis}

We implemented the algorithm on twelve classification data sets and compared the training time, testing time and accuracy with OVO and OVA. Glass, Iris, Letter, Mfeat-fac, Mfeat-fou, Mfeat-kar and Mfeat-mor, Pendigits, Satimage, Segment, Shuttle, Vowel are UCI data sets and taken from https://archive.ics.uci.edu/ml/datasets.html. The detailed properties of the data sets are given in Table I. For Letter, Shuttle and Satimage dataset, a separate training and testing file was used. But, for the rest of all the data sets 2/3 part of data is considered for training and 1/3 is considered for testing. All the datasets were scaled to [0,1] and have been randomized (similar class labels data will not appear together).
In our implementation, we make use of Radial Basis Function(RBF) Kernel and the range of value of 
 $ \gamma $ $\in$ ${2^{-10}}$ to ${2^{4}}$ , C $\in$ ${2^{-2}}$ to ${2^{12}}$. RBF kernel works best for any kind of problem [16]. In all the three approaches OVO, OVA, CBTS the best combination of $ \gamma $ and C is chosen from the above mentioned range so as to provide the higher accuracy and lower testing and training time [17]. All the computations are done on a computer with 1.60GHz Intel i5-4200U Dual core processor and RAM of 6GB and using the software LIBSVM [17].
\par In Table II, the training time and testing time of CBTS is compared with OVO and OVA. CBTS performs very well when compared with OVA. Though the training time and testing time of CBTS is little more than OVO ,  $ \gamma $ and Cost values are justified to have a good model.

\par In Table III, the results show that the accuracy of all twelve datasets by CBTS are comparable with OVO with less cost C. A higher value of C means to choose more samples as support vectors. But limiting the support vectors i.e.limiting the value of C will use the minimum memory possible and the prediction will be faster. Even when the $ \gamma $ is compared, CBTS is choosing the intermediate values as for OVO it is too small. For a good model $ \gamma $ can't be too small or too large. If $ \gamma $ is too small the model is restricted and cannot capture the complexity or “shape” of the data. If $ \gamma $ is too large there is a possibility that model will become over fit the data [18]. Coming to the comparison with OVA, CBTS gives a better accuracy.

\par In Table IV, the number of binary classifiers required for one classification is shown for OVO, OVA and CBTS. If we observe CBTS requires the minimum number of binary classifiers when compared with OVO and OVA.

\par In addition to above data sets we analyzed the Sloan Digital Sky Survey (SDSS) data set also. SDSS is a major multi-filter imaging and spectroscopic redshift survey using a dedicated 2.5-m wide-angle optical telescope at Apache Point Observatory in New Mexico, United States, astronomical telescope survey. It has 6 class labels and only 5 features (u,g,r,i,z) are considered (it has many). The data can be downloaded from SQL interface on http://skyserver.sdss.org/dr7/en/tools/search/sql.asp
\par The results are shown in table V for SDSS data set. CBTS outperforms both OVA and OVO in  the aspects of accuracy and training time with the best Gamma and Cost values. The size of the instances are varied from 30,000 to 75,000. OVA couldn't build the binary classifiers when the data size is 75,000 because of scalability issue whereas OVO and CBTS could build the required models with out any problem. Hence CBTS is scalable. And it gives a better accuracy with reduced training time. 

\begin{table}
\caption{The complete details of the data sets used in our implementation.}
\centering
\begin{tabular}{|c|c|c|c|}
\hline
Dataset & Features & Instances & Class Labels\\
\hline
Glass & 9 & 214 & 6 \\
\hline
Iris & 4 & 150 & 3 \\
\hline
Letter & 16 & 20000 & 26 \\
\hline
Mfeat-Fac & 216 & 2000 & 10 \\
\hline
Mfeat-Fou & 76 & 2000 & 10 \\
\hline
Mfeat-Kar & 64 & 2000 & 10 \\
\hline
Mfeat-Mor & 6 & 2000 & 10 \\
\hline 
Pendigits & 16 & 10992 & 9 \\
\hline
Satimage & 36 & 6435 & 6 \\
\hline
Segment & 19 & 2310 & 7 \\
\hline
Shuttle & 9 & 214 & 6 \\
\hline
Vowel & 10 & 640 & 10 \\
\hline
\end{tabular}
\end{table}

\begin{table}
\caption{Comparison of Training Time and Testing Time}
\centering
\begin{tabular}{|c|c|c|c|c|c|c|}
\hline
 Dataset &\multicolumn{2}{c|}{CBTS}&\multicolumn{2}{c|}{OVO}&\multicolumn{2}{c|}{OVA}\\
\cline{2-7}
 &Train&Test&Train&Test&Train&Test\\
\hline
Glass & 0.009 & 0.004 & 0.0028 & 0.0012 &0.025&0.013\\
\hline
Iris & 0.003 & 0.0017 & 0.001 & 0.0008 & 0.0072 & 0.0061\\
\hline
Letter & 0.535 & 0.2595 & 5.268 & 4.281&10.23&2.76\\
\hline
Mfeat-Fac & 1.0079 & 0.5077 & 0.3246 & 0.2122& 1.656&0.729\\
\hline
Mfeat-Fou & 0.6857 & 0.3179 & 0.2591 & 0.1624&0.962&0.423 \\
\hline
Mfeat-Kar & 0.4391 & 0.2221 & 0.1491 & 0.0887 &1.244&0.552\\
\hline
Mfeat-Mor & 0.0823 & 0.0545 & 0.0652 & 0.0349&0.4773&0.0726 \\
\hline 
Pendigits & 0.794 & 0.657 & 0.225 & 0.175 & 0.7748 &0.2398\\
\hline
Satimage & 0.535 & 0.2595 & 0.2596 & 0.173 &1.006&0.3503\\
\hline
Segment & 0.125 & 0.071 & 0.078 & 0.034 &0.244&0.0669 \\
\hline
Shuttle & 3.9367 &9.0613 & 2.221 & 0.394 & 4.491 & 0.606\\
\hline
Vowel & 0.028 & 0.0131 & 0.0131 & 0.006 & 0.057 & 0.032 \\
\hline
\end{tabular}
\end{table}

\begin{table*}
\caption{Comparison of $\gamma$ , Cost(C) and Accuracy.}
\centering
\begin{tabular}{|c|c|c|c|c|c|c|c|c|c|}
\hline
 Dataset &\multicolumn{3}{c|}{CBTS}&\multicolumn{3}{c|}{OVO}&\multicolumn{3}{c|}{OVA}\\
\cline{2-10}
 &Gamma&Cost&Accuracy&Gamma&Cost&Accuracy&Gamma&Cost&Accuracy\\
\hline
Glass&$2^{4}$&$2^{-1}$&73.61&$2^{-3}$&$2^{5}$&76.38 & $2^{-1}$&$2^{9}$ & 68.49\\
\hline
Iris&$2^{4}$&$2^{-1}$&98&$2^{-10}$&$2^{12}$&98 & $2^{1}$&$2^{2}$ &98.03 \\
\hline
Letter & $2^{4}$ & $2^{0}$ & 93.68 & $2^{1}$ & $2^{4}$ & 96.68 & $2^{0}$&$2^{10}$ &91.88 \\
\hline
Mfeat-Fac & $2^{2}$ & $2^{-2}$ & 97.75 & $2^{-8}$ & $2^{11}$ & 98.2 & $2^{-3}$&$2^{12}$ &97.15\\
\hline
Mfeat-Fou & $2^{4}$ & $2^{-2}$ & 82.46 &$2^{-4}$ & $2^{12}$ & 85.75 & $2^{-4}$&$2^{3}$ &80.08\\
\hline
Mfeat-Kar & $2^{4}$ & $2^{-2}$ & 98.05 & $2^{-10}$ & $2^{9}$ & 97.00 & $2^{-3}$&$2^{12}$ &96.25 \\
\hline
Mfeat-Mor & $2^{4}$ & $2^{-2}$ & 72.26 & $2^{-6}$ & $2^{12}$ & 73.31 & $2^{-1}$&$2^{10}$ &68.41 \\
\hline 
Pendigits & $2^{4}$ & $2^{-2}$ & 99.67 & $2^{-4}$ & $2^{6}$ & 99.67 & $2^{-3}$&$2^{5}$ &99.31\\
\hline
Satimage & $2^{3}$ & $2^{-2}$ & 88.35 & $2^{-3}$ & $2^{3}$ & 88.50 & $2^{-2}$&$2^{2}$ &84.82\\
\hline
Segment & $2^{4}$ & $2^{0}$ & 96.88& $2^{-1}$ & $2^{12}$ & 97.66 & $2^{-1}$&$2^{11}$&95.33\\
\hline
Shuttle & $2^{4}$ & $2^{5}$ & 99.93 & $2^{4}$ & $2^{11}$ & 99.99& $2^{3}$&$2^{11}$ &99.96\\
\hline
Vowel & $2^{2}$ & $2^{1}$ & 97.15 & $2^{-1}$ & $2^{6}$ & 98.29& $2^{0}$&$2^{5}$ &93.22\\
\hline 
\end{tabular}
\end{table*}

\begin{table}
\caption{The Number of binary SVMs required for one classification.}
\centering
\begin{tabular}{|c|c|c||c|}
\hline
Dataset & OVO & OVA & CBTS \\
\hline
Glass & 36 & 9  & 8 \\
\hline
Iris & 6 & 4 & 3 \\
\hline
Letter & 120 & 16 & 15 \\
\hline
Mfeat-Fac & 23220 & 216 & 215 \\
\hline
Mfeat-Fou & 2850 & 76 & 75  \\
\hline
Mfeat-Kar & 2016 & 64 & 63  \\
\hline
Mfeat-Mor & 15 & 6 & 5 \\
\hline 
Pendigits & 120 & 16 & 15  \\
\hline
Satimage & 630 & 36 & 35  \\
\hline
Segment & 171 & 19 & 18  \\
\hline
Shuttle & 36 & 9 & 8 \\
\hline
Vowel & 45 & 10 & 9  \\
\hline
\end{tabular}
\end{table}

\begin{table}
\caption{Large (SDSS) data set Analysis }
\centering
\begin{tabular}{|c|c|c|c|c|c|c|}
\hline
SDSS & Algorithm & Accuracy & Gamma & Cost &Tr.Time &Te.Time \\
\hline
 &OVA & 80.42 & $2^{4}$ & $2^{-2}$ & 19.05 & 4.78 \\
\cline{2-7}
 30,000 &OVO & 86.61 & $2^{1}$ & $2^{6}$ & 11.37 & 3.23 \\
\cline{2-7}
 &CBTS & 86.75  & $2^{1}$ & $2^{6}$ & 10.65 & 6.24 \\     
\hline

&OVA & 85.06 & $2^{4}$ & $2^{11}$  & 785.31 & 9.57 \\
\cline{2-7}
 50,000 &OVO & 87.08 & $2^{4}$ & $2^{6}$ & 34.99 & 9.94 \\
\cline{2-7}
 &CBTS & 87.13 & $2^{1}$ & $2^{6}$ & 33.18 & 22.99 \\     
\hline

& OVA & X & X & X & X & X \\
\cline{2-7}
75,000 & OVO & 86.98 & $2^{3}$ & $2^{0}$ & 67.96 & 32.32 \\
\cline{2-7}
& CBTS & 86.26 & $2^{1}$ & $2^{2}$ & 60.59 & 49.95 \\
\hline
\end{tabular}
\end{table}

\section{Conclusion}
We propose a new algorithm CBTS (Centroid based Binary Tree Structured SVM) , a binary tree structure in which the root node contains all the class labels and is splitted based on the K-means Clustering at the first step. And then the left and right nodes are recursively splitted into half till we have only two class labels. In this N-1 SVMs are required to construct  but the required SVMs in OVO,OVA are N(N-1)/2 , N  respectively. The testing time for all OVO, OVA, DAG is linear on N i.e. O(N) but for our algorithm it is O(log N) as it needs only (N-1)/2 classifiers to predict the class label for the test object. Experimental results show that the accuracy of CBTS is comparable with OVO approach and it out performs with OVA both in accuracy and training,testing time. CBTS also capable of handling large data sets, hence scalable.
By using a more optimized approach of implementing CBTS, testing time may further be reduced. Instead of K-Means clustering alternative partition techniques can also be explored at the root level. Furthermore, a parallel/distributed version of our algorithm can be taken up as a future work. In our algorithm there are independent SVM models that can be constructed simultaneously which results in reducing both training and testing time of CBTS even further. 


\section*{Acknowledgment}

We are thankful for the resources provided by the Department of Computer Science and Information Systems, BITS, Pilani, K.K. Birla
Goa Campus to carry out the experimental analysis.




\begin{thebibliography}{1}

\bibitem{IEEEhowto:kopka}
C.Cortes and V. Vapnik. \emph{Support Vector network. Machine Learning}.
\hskip 1em plus
0.5em minus 0.4em\relax
September 1995, Volume 20, Issue 3, pp 273-297
\bibitem{IEEEhowto:kopka}
V. Vapnik. \emph{The Nature of Statistical Learning
Theory}. NY: Springer-Verlag. 1995. 
\bibitem{IEEEhowto:kopka}
K.Crammer,Y.Singer, \emph{On the learnability and design of output codes for multiclass problems}.
\hskip 1em plus
0.5em minus 0.4em\relax
Proceedings of the Thirteenth Annual Conference on Computational Learning Theory, Morgan Kaufmann Publishers Inc., San Francisco,CA,USA,2000,pp.35-46
\bibitem{IEEEhowto:kopka}
V.N.Vapnik \emph{Statistical Learning Theory}.
\hskip 1em plus
0.5em minus 0.4em\relax John Wiley and Sons,New York,1998.
\bibitem{IEEEhowto:kopka}
Y.Guermeur, Combining discriminant models with new-multi-class SVMs,
\hskip 1em plus
0.5em minus 0.4em\relax
Patt.Anal.Appl.5(2002) 168-179
\bibitem{IEEEhowto:kopka}
Chih-Wei Hsu and Chih-jen Lin. \emph{A comparison of Methods for Multi Class Support Vector Machines}.
\hskip 1em plus
0.5em minus 0.4em\relax
IEEE Transactions on Neural Networks, VOL. 13, NO. 2, March 2002
\bibitem{IEEEhowto:kopka}
Abe, Shigeo \emph { Analysis of Multi Support Vector Machines}.
\hskip 1em plus
0.5em minus 0.4em\relax
Proc. International Conference on Computational Intelligence for Modelling Control and Automation
(CIMCA’2003), : 385-396
\bibitem{IEEEhowto:kopka}
Ryan Rifkin ,Aldebano Klautau, \emph { In Defense of One--Vs-All Classification }
\hskip 1em plus
0.5em minus 0.4em\relax
Journal of Machine Learning Research 5 (2004) 101-141.
\bibitem{IEEEhowto:kopka}
Friedman. \emph {Another approach to polychotomous classification. Technical report}, 
\hskip 1em plus
0.5em minus 0.4em\relax 
Stanford University, 1996.
\bibitem{IEEEhowto:kopka}
Trevor Hastie and Robert Tibshirani. \emph {Classification by pairwise coupling.} 
\hskip 1em plus
0.5em minus 0.4em\relax 
Advances in Neural Information Processing Systems, volume 10. The MIT Press,1998
\bibitem{IEEEhowto:kopka}
Erin Allwein, Robert Shapire, and Yoram Singer. \emph{Reducing multiclass to
binary: A unifying approach for margin classifiers}.
\hskip 1em plus
0.5em minus 0.4em\relax 
Journal of Machine Learning Research, pages 113–141, 2000

\bibitem{IEEEhowto:kopka}
L.Bottou, C. Cortes, J. Denker, H. Drucker, I. Guyon, L. Jackel, Y. LeCun, U. Muller, E. Sackinger, P. Simard, and V. Vapnik. \emph{Comparison of classifier methods : a case study in handwriting digit recognition}.
\hskip 1em plus
0.5em minus 0.4em\relax
In International Conference on Pattern Recognition, pages 77-87. IEEE Computer Society Press, 1994.
\bibitem{IEEEhowto:kopka}
S. Knerr, L. Personnaz, and G. Dreyfus, \emph{Single-layer learning revisited: a stepwise procedure for building and training a neural network}.
\hskip 1em plus
0.5em minus 0.4em\relax 
Neurocomputing: Algorithms, Architectures and Applications. Springer-Verlag, 1990.
\bibitem{IEEEhowto:kopka}
J.Friedman. \emph{Another approach to polychotomous classification. Technical report, Department of Statistics, Stanford University,1996}.
\hskip 1em plus
 0.5em minus 0.4em\relax Available at http://www-stat.stanford.edu/reports/friedman/poly.ps.Z
\bibitem{IEEEhowto:kopka}
U. Krebel. \emph{Pairwise classification and support vector machines}.
\hskip 1em plus
0.5em minus 0.4em\relax Advances in Kernel Methods – Support Vector learning, pages 255-268, Cambrdige, MA, 1999. MIT Press.

\bibitem{IEEEhowto:kopka}
K.P.Soman, R.Loganathan, and V.Ajay. \emph{Machine Learning with SVM and Other Kernel Methods}.
\hskip 1em plus
0.5em minus 0.4em\relax 
PHI, pages 153-155, 2011.

\bibitem{IEEEhowto:kopka}
Chih-Chung Chang and Chih-Jen Lin .\emph{LIBSVM : a library for support vector machines.}
\hskip 1em plus
0.5em minus 0.4em\relax 
ACM Transactions on Intelligent Systems and Technology, 2:27:1--27:27, 2011. Software available at http://www.csie.ntu.edu.tw/~cjlin/libsvm 

\bibitem{IEEEhowto:kopka}
Buitinck et al. \emph {API design for machine learning software: experiences from the scikit-learn project}. 2013.
\hskip 1em plus 
0.5em minus 0.4em\relax Available at $http://scikit-learn.org/stable/auto_examples/svm/plot_rbf_parameters. \\
html$

\end{thebibliography}
%

\end{document}